# A Study of the Effect of Resolving Negation and Sentiment Analysis in Recognizing Text Entailment for Arabic

Fatima T. AL-Khawaldeh
Department of Computer Science, Al-Albayt University
Al-Mafraq, Jordan

Abstract— Recognizing the entailment relation showed that its influence to extract the semantic inferences in wide-ranging natural language processing domains (text summarization, question answering, etc.) and enhanced the results of their output. For Arabic language, few attempts concerns with Arabic entailment problem. This paper aims to increase the entailment accuracy for Arabic texts by resolving negation of the text-hypothesis pair and determining the polarity of the text-hypothesis pair whether it is Positive, Negative or Neutral. It is noticed that the absence of negation detection feature gives inaccurate results when detecting the entailment relation since the negation revers the truth. The negation words are considered stop words and removed from the text-hypothesis pair which may lead wrong entailment decision. Another case not solved previously, it is impossible that the positive text entails negative text and vice versa. In this paper, in order to classify the text-hypothesis pair polarity, a sentiment analysis tool is used. We show that analyzing the polarity of the text-hypothesis pair increases the entailment accuracy. to evaluate our approach we used a dataset for Arabic textual entailment (ArbTEDS) consisted of 618 text-hypothesis pairs and showed that the Arabic entailment accuracy is increased by resolving negation for entailment relation and analyzing the polarity of the text-hypothesis pair.

## I. INTRODUCTION

Recognizing Text Entailment is a general process to capture major semantic inferences among texts. Textual Entailment is inferring a text from another. Entailment is a directional relation between two texts. The entailing fragment is called a text and entailed one is called a hypothesis [1]. Recognizing Text Entailment tasks presented in RTE challenges.

In RTE 1 (2005), the first attempt to develop entailment task that able to capture major semantic inferences in NLP applications was provided in [2]. Low accuracy achieved in RTE 1 (2005). In RTE 2 (2006), more attempts to judge entailment relation where a hypothesis H is entailed by a text T. The highest accuracy achieved in In RTE 2 (2006) was 0.7538 [3]. In RTE 3 (2007), the systems developers' share their ideas and resources to get better output. The best accuracy of 80% is achieved in [4]. In RTE 4 (2008) three way classification (Entailment, Contra-diction and Unknown) analyzed to get more precise information. Less accuracy obtained in this system retuned to 0.746%. In RTE 5 (2009), the main NLP applications the entailment task oriented to them are: Question Answering, Information Extraction, and Information Retrieval, Text Summarization. In this challenge the highest accuracy obtained was 0.6833. In RTE 6 (2010) and RTE 7 (2011) recognizing textual entailment is dedicated in two NLP application: Summarization and Knowledge Base Population settings. Since this task focuses on summarization, f-measure is used as evaluated measure .The highest F-measure was 0.4801.

The main evaluation measures used to evaluate the results of the RTE tasks are: Accuracy, Precision, and Recall. The accuracy is defined by PASCAL RTE challenge: the ratio of correct entailment decision to the total number of entailment problem. Precision is the ratio of number of correctly predicted entailment to the number of predicted entailment. Recall the ratio of number of correctly predicted entailment to the actual number of correct entailments [1].

In spite of Arabic is Semitic language of 22 countries and is spoken by more than 300 million people, very few researchers tackled text entailment problem. The main problem of previous Arabic text entailment systems is that not recognizing the negation, where the negation revers the truth and not taking the sentiment polarity into consideration where negative or positive feeling doesn't entail the opposite feeling.

Several applications of Natural Language Processing like information extraction, textual entailment and sentiment analysis concern in the handling of negation. A significant impact is obtained when treatment of negation on NLP applications [5]. For example, in sentiment analysis, the polarity of the statement should be the opposite of its negation [5].

In [20], the authors proved that, for English language, good precision is obtained to detect the entailment and non-entailment relation by sentiment analysis of the text-hypothesis pair. Sentiment analysis is the task of classifying a text according to its polarity positive and negative opinions, using natural language processing and computational techniques. For English language, Different approaches have been followed to analyze sentiment. Less work in Sentiment analysis is done for Arabic language.

To our knowledge, this work represents the first attempt to notice the influence of resolving the negation and sentiment polarity in the Recognition of Textual Entailment for Arabic language. We conducted an experiment with Arabic Text Entailment Dataset (ArbTEDS) [21]; showing that more entailment accuracy obtained by resolving the negation and discovering sentiment polarity of the text-hypothesis pair. In this work, we tackle the problem of developing an Arabic text entailment system that produces more accurate results.

This paper is organized in 7 sections. Related works are illustrated in section 2. The Arabic Text Entailment phases (ATE) are summarized in section 3. Resolving Negation in Recognizing Text Entailment for Arabic is illustrated in section 4. Sentiment Analysis adaptation for Recognizing Arabic Text Entailment is discussed in section 5. The experimental results are discussed in section 6. Finally, we concluded in section 7.

## II. RELATED WORKS

For Arabic language, few attempts concern about studying RTE. In 2011, Arabic Textual Entailment system called (ArbTE) is developed in [6]. ArbTE examined the effectiveness of existing techniques for textual entailment when applied to Arabic language. In this research, the basic version of the [7] TED algorithm was extended and inference rules were applied to obtain more accurate results. It is found that it is effective to combine or apply some of existing techniques for textual entailment to Arabic. The first dataset for a text entailment for Arabic was published in [8] and publically available. The dataset is called (ArbTEDS), consists of 618 T-H pairs. Two tools were used to automatically collecting T-H pairs from news websites and to annotate pair which collected by hand [8]. In previous studies, it was found that implementing entailment algorithm suggested by [9] with some modification suited to Arabic texts, it improved the performance of Arabic text summarization systems [10] and Arabic why question answering systems [11]. The effectiveness of ATE was evaluated by measuring its influence on the output of summarization and question answering.

For sentiment analysis, Researchers have suggested many different approaches. In the work of the authors of [12], Entropy Weighted Genetic Algorithms combine Genetic algorithms is implemented to select the feature sentiment analysis that work for multiple languages. The authors of [13] used Local Grammar to extract sentiment features from financial news domain applied to Arabic, English and Chinese languages. A web search engine annotates returned pages of Arabic business reviews with sentiment analysis results is built in [14]. A combined classification approach of a lexicon-based classifier and a maximum entropy classifier is proposed in [15] for sentiment classification. The authors of [16] built a new sentiment analysis tool called colloquial Non-Standard Arabic - Modern Standard Arabic-Sentiment Analysis Tool (CNSAMSA-SAT). One of the main tasks of this tool is building polarity lexicons oriented to both colloquial Arabic and MSA. In 2015, the authors of [17] generated large multi-domain datasets for sentiment analysis in Arabic and built multi-domain lexicons from the produced datasets are publically available in [18].

## III. THE ARABIC TEXT ENTAILMENT PHASES (ATE)

The entailment algorithm which is implemented in [10] and [11] will be referred as Arabic Text Entailment (ATE) in this document. The proposed system in this paper called Sentiment Analysis and Negation Resolving for Arabic Text Entailment (SANATE)

The main phases of ATE are:

- Removing stop words (irrelevant words).
- Word Stemming.
- Extracting the related words for each word in the text-hypothesis pair using Arabic WordNet. The related words are obtained by extracting all the possible senses (words semantically related).
- Calculating the common words (c) in the text-hypothesis pair where common words are words have the same roots or words related by semantic relations.
- Determining the length of text-hypothesis pair such the length of text is m and the length of hypothesis is n.
- Verifying that m ≥ n ≥ c.

- Appling the three methods equations are used by [9]:

$cosT(T;H) = \sqrt{c/m}$ …………………….(1)
$cosH(T;H) = \sqrt{c/n}$ …………………..(2)
$cosH \cup T(T;H) = \sqrt{4c^2/(n+c)(m+c)}$ ….(3)
- Satisfying this primary condition $cosH(T,H) \geq cosH \cup T(T,H) \geq cosT(T,H)$

- Verifying the compulsory conditions to satisfy the entailment relation. The compulsory conditions are:

cosH∪T−cosT≤τ1 ..................... (4)
cosH - cosH∪T≤τ ......................(5)
Max {cosT;cosH;cosH∪T}≥τ3 ......(6)

Experimentally, the thresholds used are: τ1=0.095, τ2=0.2, τ3=0.5. The decision of entailment is: entails, if all conditions checked successfully and not entails if one or more condition not satisfied. SANATE system is shown in Figure 1.

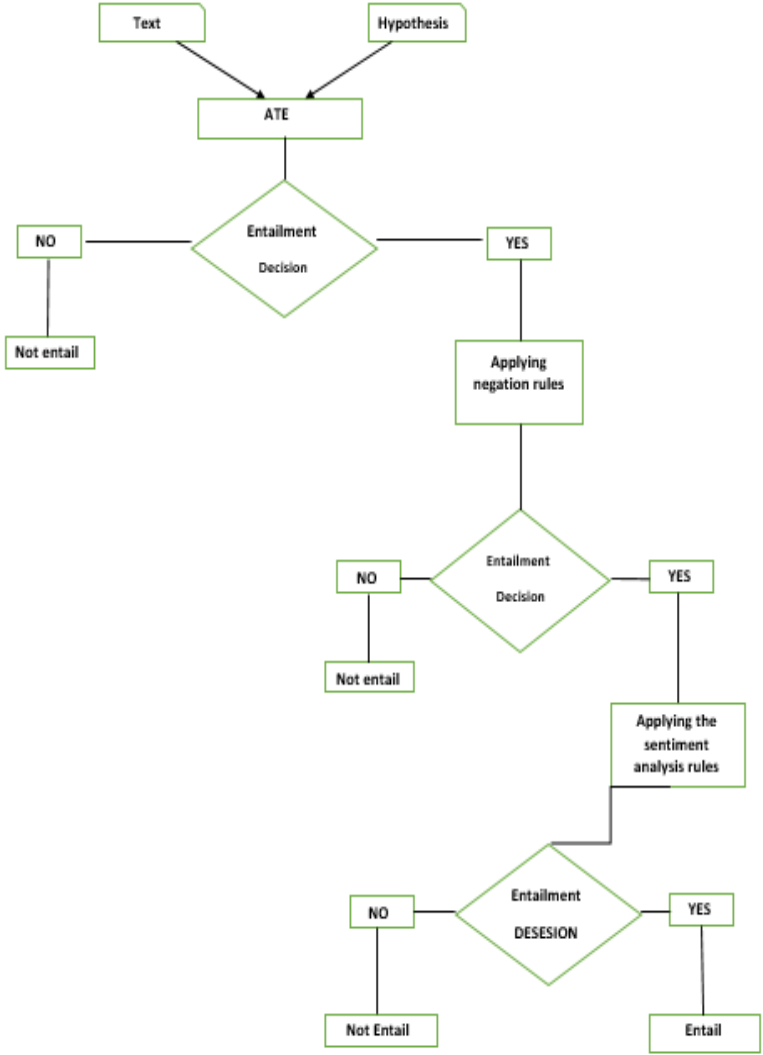

Figure 1: General diagram of SANATE system

## IV. RESOLVING NEGATION IN RECOGNIZING TEXT ENTAILMENT FOR ARABIC

It is noticed that ATE algorithm didn't take the negation into consideration which may lead less accurate results. Negation reverse the value of truth, for example, suppose that we have the text-hypothesis pair (T, H):

T: انا احب قراءة الكتب.

H: انا لا احب قراءة الكتب

The fact that in H: T is negated by the negation particle (لا) which is removed as one of the words listed in stop words list. According to ATE, the entailment decision is T entails which is wrong decision. The inaccuracy comes from not recognizing the negation when entailment relation is checked.

Five different Arabic negative particles are debated in [19]: particle *maa* (ما), the particle *laa* (لا) , the particle *lam* (لم) ,the particle *lan* (لن), and the *particle laysa* (ليس).

For improving ATE, we put some of rules to detect the negation. If the entailment decision by the ATE is "not entail" then no another checking to entailment is done. If the entailment decision by the ATE is "entail" then a set of rules will be checked. The rules applied to make another checking to text and hypothesis entailment are:

1) If the negative particle appears before the same verb (common verb) in the text or hypothesis, then the judgment is: not entails, for example:

T: انا لم اقرأ الجريدة
H: انا اقرأ الجريدة

The negative particle *lam* (لم) appears before the common verb ((اقرا)), the entailment judgment is not entails.

2) If the negative particle appears before the common verb in the text and hypothesis (both), then the judgment is: entails, for example:

T : ليس يعلم الغيب الا الله
H: لا يدرك الغيب الا الله

Negative particle *laysa* ليس appears before the common verb (يعلم) in T and negative particle *laa* (لا) appears before the common verb (يدرك). Notice that (يدركand (يعلم) are semantically related.

3) If the T and H have different verb and the negative particle appears before a verb in the text or hypothesis or both, then the judgment is: NOT entails, for example:

T: لم اقرأ الجريدة
H: انا لا اشتري الجريدة

The negative particle *lam* (لم) appears before a word (اقرا) and the negative particle *laa* (لا) *appears* before another verb (اشتري) which are not common.

4) If text-hypothesis pair have more than verbs (common verbs), then:
   a) If one of the common verbs is negated by one of the negative particles in T or H but not

negated in both (T and H), the judgment is not entail. If the common verb is negated by one of the negative particles in T and H (both), the judgment is entail.

## V. SENTIMENT ANALYSIS adaptation in Recognizing Arabic Text Entailment

Another enhancement to ATE algorithm will be done by polarity analysis. In this paper (CNSAMSA-SAT) Tool is used to automatically identify the polarity (Positive/Negative/Neutral) of text-hypothesis pair, since the accuracy of this tool reached to 90% and covers eight domains: Books, Movies, Places, Politics, Products, Social, Technological, and Educational. We used in SANATE, Multi domain lexicon files generated in [17]. The authors of [17] showed that the generated lexicons they built are effectiveness and reliable for Arabic sentiment analysis. Multi domain lexicon files are available in [18]. The combined lexicon (ALL_lex.csv) includes hotel, library, movie, production and restaurant opinion words with its polarity. We split this file into two file: Positive Sentiment Dictionary and Negative Sentiment Dictionary, In order to use them in the: CNSA-MSA-SAT Algorithm.

**Algorithm:** CNSA-MSA-SAT.
**Input:**
***R*:** text-hypothesis pair
***T*:** the set of the Opinion words (ALL_lex.csv)
***PD*:** the set of Positive Sentiment Dictionary
***ND*:** the set of Negative Sentiment Dictionary
**Output:**
  *R1={P, N, NT}, where P: Positive, N: Negative, NT: Neutral*

  Begin

  1. *For each ti $\in$ T do*
  2. *Search for ti in PD where ti $\in$ T*
  3. *If ti $\in$ PD then*
  4. *Pos-TF= Pos-TF +1*
  5. *Else*
  6. *Search for ti in ND where ti $\in$ T*
  7. *If ti $\in$ ND then*
  8. *Neg-TF = Neg-TF +1*
  9. *End For*
  10. *If (Pos-TF $\geq$ 2) && (Pos-TF > Neg-TF) then*
  11. *R1= P*
  12. *End If*
  13. *If (Neg-TF $\geq$ 2) && (Pos-TF < Neg-TF) then*
  14. *R1 = N*
  15. *End If*
  16. *If (Pos-TF == Neg-TF) && (Pos-TF != 0)then*
  17. *R1=NT*
  18. *Return R1*
  19. *End If*
  20. *End*

5) If text and hypothesis have Opinion words, as in the following example ارى is opinion word then:
   a) If the output R dissimilar in T and H which means different polarity, judgment is not entail, for example:

T: الى الان لم ارى نتيجة جيده ----------Negative opinion
H: من النظره الاولى ارى انها تؤدي المطلوب ------- Positive opinion

   b) If the output R similar in T and H, judgment is entail, for example:

T: الى الان لم ارى نتيجة جيده --------Negative opinion
H: من النظره الاولى ارى انها لا تؤدي المطلوب ---------
Negative opinion

## VI. THE EXPERIMENTAL RESULTS

In order to evaluate our approach we used a dataset for Arabic textual entailment (ArbTEDS) consisted of 618 text-hypothesis pairs. Each text-hypothesis pair of ArbTEDS is entered to ATE system and SANATE system. The accuracy of each system is calculated by accuracy equation: the ratio of correct entailment decision to the total number of entailment problem. The accuracy of ATE is 0.617 and the accuracy of SANATE is 0.693.

From the results of accuracies of ATE system and SANATE system, it is shown that resolving the negation and classifying the text to its polarity by sentiment analysis, increases the performance of detecting the entailment relation and non-entailment relation. Figure 2 shows Comparison of accuracies of ATE and SANATE systems. It is clearly that SANATE performance is better than ATE which illustrates the influence of representing the negation and sentiment analysis for recognizing the Arabic text entailment.

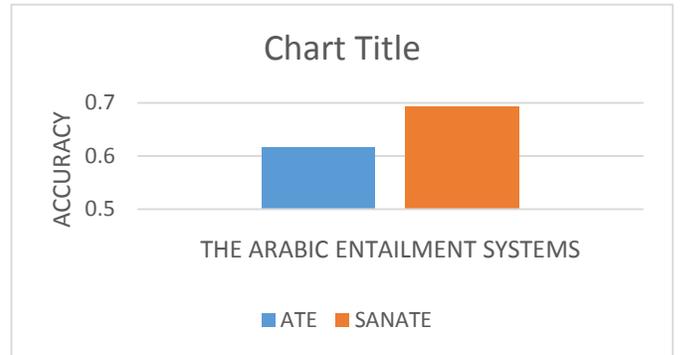

Figure 2: Comparison of accuracies of ATE and SANATE

## VII. CONCLUSION AND FUTURE RESEARCH

We conducted an experiment with a dataset for Arabic textual entailment (ArbTEDS); showing that it is more accurate results to detect entailment by representing the negation and analyzing the polarity of the text-hypothesis pair.

Without resolving the negation entailment relation decision may be the opposite, since the negation gives the opposite of truth. Some texts may entail hypothesis but the existing of negative particles reverse the judgment of entailment relation. Another result from our experiments, detecting the polarity of text and hypothesis pair impacts significantly on detecting entailment relation and non-entailment relation. It is impossible positive opinion entails negative opinion and vice versa.